\documentclass[doublecol,linenumbers]{epl2} 

\usepackage{amsmath}
\usepackage{amsfonts}
\usepackage{hyperref}

\usepackage{color}
\definecolor{redcolor}{rgb}{1.0,0.,0.}

\title{Concentric network symmetry grasps authors' styles in word adjacency networks}
\shorttitle{Network symmetry reveals authors' styles in word adjacency networks} 

\author{Diego R. Amancio\inst{1} \and Filipi N. Silva\inst{2} \and Luciano da F. Costa\inst{2}}

\institute{
  \inst{1} Institute of Mathematics and Computer Science\\
           \ \ \ \ \ \ University of S\~ao Paulo, S\~ao Carlos, S\~ao Paulo, Brazil \\

  \inst{2} S\~ao Carlos Institute of Physics \\
           University of S\~ao Paulo, S\~ao Carlos, S\~ao Paulo, Brazil \\
}


\abstract{
Several characteristics of written texts have been inferred from statistical analysis derived from networked models. Even though many network measurements have been adapted to study textual properties at several levels of complexity,
some textual aspects have been disregarded. In this paper, we study the symmetry of word adjacency networks, a well-known representation of text as a graph. A statistical analysis of the symmetry distribution performed in several novels showed that most of the words do not display symmetric patterns of connectivity. More specifically, the merged symmetry displayed a distribution similar to the ubiquitous power-law distribution. Our experiments also revealed that the studied metrics do not correlate with other traditional network measurements, such as the degree or betweenness centrality.
The effectiveness of the symmetry measurements was verified in the authorship attribution task. Interestingly, we found that specific authors prefer particular types of symmetric motifs. As a consequence, the authorship of books could be accurately identified in $82.5\%$ of the cases, in a dataset comprising books written by $8$ authors. Because the proposed measurements for text analysis are complementary to the traditional approach, they can be used to improve the characterization of text networks, which might be useful for applications such as identification of topical words and information retrieval.
}

\begin{document}

\maketitle

\section{Introduction}

In recent years, network science has become {commonplace}. Many real systems such as the Internet, social networks and transportation systems have increasingly been studied via networked models~\cite{surveyApp}. {Because language is organized by rules and relationships between words in a complex way, it can also be represented as networks}. In this case, words are connected according to syntactical or semantical relationships~\cite{cong}. The use of the network framework not only allowed for a better understanding of the origins and organization of language~\cite{cong}, but also improved the performance of several natural processing language tasks, including e.g. the automatic summarization of texts~\cite{extractive}, the identification of word senses~\cite{wsd} and the classification of syntactical complexity~\cite{compSanda}.

Many measurements proposed for analyzing complex networks have been reinterpreted when applied to analyze linguistic features. Centrality measurements, for example, have been useful to identify core concepts and keywords, which in turn have allowed the improvement of summarization and classification tasks~\cite{extractive}. While a myriad of measurements have been adapted to probe textual patterns, only a limited number of studies have been devoted to devise novel network measurements that are able to identify more complex linguistic patterns. Particularly, a relevant pattern that has not been addressed by networked-linguistic models is the quantification of the heterogeneity of specific textual distributions. This is the case of the spatial distribution of words along the text, which has been mainly studied in terms of the \emph{burstiness} (or \emph{intermittency}) of time series~\cite{ortuno}. Another interesting pattern concerns the uneven distribution of the number of distinct neighbors of words~\cite{nova}.
In this context, we introduce two network measurements to quantify the heterogeneity of accessing words neighbors in word adjacency networks. As we shall show, the adopted measurements, henceforth referred to as symmetry measurements, are able to characterize authors' stylistic marks, since distinct authors display specific bias towards particular network motifs. In addition to being useful to improve the characterization of word adjacency networks, we found out that the symmetry measurements do not correlate with other traditional network measurements. Therefore, they could be useful to complement the characterization of text networks in its several levels of complexity.
%

\section{Methods}

In this section, we describe the formation word adjacency networks from raw books. The symmetry measurements, namely backbone and merged symmetry are then described. Furthermore, we present a short introduction to the pattern recognition methods employed in this study.

\subsection{Word adjacency networks}

Written texts can be modeled as networks in several ways~\cite{cong}. If one aims at grasping stylistic textual features, networks generated from syntactical analysis can be employed~\cite{patterns,extractive}. Another possibility is to map texts into a word adjacency network (WAN), which links adjacent words~\cite{literary,short,poetry}.  Actually, WANs can be considered as an extension of the syntactical model since most of the syntactical links occur between adjacent words~\cite{patterns}. Because syntax depends upon the language, WANs have also proven useful to capture language dependent features~\cite{voynich}.

To construct a word adjacency network, some pre-processing steps are usually applied. First, \emph{stopwords} such as articles and prepositions are removed because such words convey no semantic information. Therefore, they can be modeled as edges in the WAN model because \emph{stopwords} usually play the role of linking content words. In order to represent as a single node the words that refer to the same concept, the text undergoes a lemmatization process. Hence, nouns and verbs are mapped to their singular and infinitive forms, respectively. To minimize the errors arising from the lemmatization, before this step, all words are labeled with their part-of-speech~\cite{statnlp}. In the current study, the maximum-entropy model devised in~\cite{ratna} was used to perform the part-of-speech labeling. After the lemmatization and the removal of the \emph{stopwords}, each distinct word is mapped into a node and edges are created between adjacent words. Further details regarding the WAN model can be found in~\cite{literary}.


\subsection{Symmetry in networks}
\begin{figure*}
\onefigure[width=0.90\linewidth]{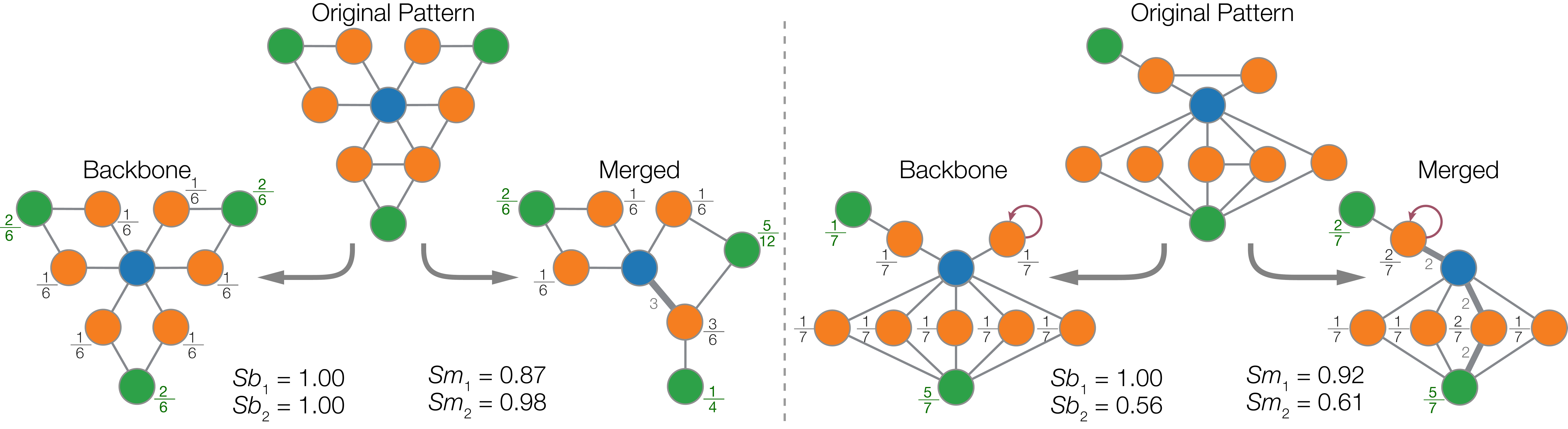}
\caption{Example illustrating the calculation of the backbone and merged symmetries for two concentric $2$-patterns. The numbers next to each node account for the transition probabilities and colors indicate the respective concentric level of a node (blue for level $0$, orange for level $1$ and green for level $2$)~\cite{concentric}. Red self loops indicate a dead end. Both transformations of patterns are shown. The backbone pattern is obtained by removing edges connecting nodes at the same level from the original pattern, whereas merged patterns are weighted subgraphs created by merging nodes originally connected at the same concentric level. In this case, the weight corresponds to the number of connections spanning from the nodes that were merged to each node in other concentric levels. Note that the pattern in the left panel only presents merged asymmetry, while the pattern in the right panel presents both types of asymmetry, which is also confirmed by the symmetry values.}
\label{fig.symmetry}
\end{figure*}
Symmetry is one of the most fundamental aspects of complex systems, naturally emerging from physical spatial restrictions and laws~\cite{sym1}, self organization~\cite{sym2}, biological structures~\cite{sym3}, and chemical reactions~\cite{sym4}, etc. Written texts bear no exception to this rule, presenting intrinsic patterns of symmetry. In a sentence, for instance, some words can be exchanged by synonyms without compromising its original meaning. In a similar fashion, some grammatical constructions are also interchangeable. Aside from restrictions conveying semantic relationships and grammatical rules, authors also tend to employ additional restrictions in their works, which in turn affects their written style. 
Whenever texts are represented by networks, it is expected that such styles may be reflected on the symmetrical characteristics of its topological structure.

While the concept of symmetry in graph theory is tightly related to the problem of finding and counting automorphisms, this approach cannot be straightforwardly extended to study most of real complex networks~\cite{holme}. Recently, practical definitions of symmetries for real networks have been proposed in the literature. They include path similarity techniques~\cite{holme}, the methods based on quantum walks~\cite{ingleses} and concentric rings~\cite{mir}.
The latter presents some advantages over the other strategies.
For example, the symmetry can be calculated locally around nodes in a multiscale fashion, defined in terms of node centered subgraphs referred to as \emph{concentric patterns} conceptually linked to the concentric levels of a node. The concentric level $\Gamma_h(i)$ is defined as the set of nodes $h$ hops away from the original node $i$ and the concentric $l$-pattern is the subgraph comprising only nodes located $l$ or less hops away from $i$,  i.e., nodes in the set $\bigcup_{h=0}^l \Gamma_{h}(i)$.

The concentric symmetry approach is based on the accessibility measurement~\cite{viana}, which is calculated as a normalization of the entropy obtained from the transition probabilities for a network walk dynamics, such as the traditional random walk or self-avoiding random walk. In particular, the symmetry is obtained considering a very special case of walk dynamics in which an agent never goes back to a node belonging to a lower concentric level.
%
%
However, to account for the degeneracy caused by connections between nodes in the same concentric level, two transformations of concentric patterns were proposed, resulting in two types of symmetry measurements: \emph{backbone} and \emph{merged} symmetries. The backbone symmetry, $Sb$, is {loosely} based on the concept of radial symmetry, in which edges among nodes in the same concentric level are removed for the pattern. Differently, the merged symmetry, $Sm$, that bears some resemblance with angular symmetry, is obtained from patterns by effectively merging nodes in the same concentric level. In both cases, the symmetry measurements $S_h$ for level $h$ centered on $i$ are calculated from the Shannon entropy $H_h$ of the transition probabilities $P_h(i \to j)$. More specifically,
\begin{equation} \label{equation:symmetry}
S_h(i) = \frac{\exp{\left\{\sum\limits_{j\,\in\,\Gamma_{h}(i)} P_h(i \to j)\ln[P_h(i \to j)]\right\}}}{|\Gamma_{h}(i)| + \sum_{r=0}^{h-1} \Xi_r},
\end{equation}
where $\Xi_r$ stands for the number of dead ends in level $r$ (i.e., nodes with no connections to any node in the next concentric level). Fig.~\ref{fig.symmetry} illustrates the backbone and merged transformations for two patterns alongside the transition probabilities and calculated symmetries.

\subsection{Pattern Recognition Methods}

Pattern recognition methods are useful to identify patterns and infer classifiers~\cite{duda}. Particularly, in this study, pattern recognition methods were applied to recognize patterns in the distribution of symmetry measurements across distinct authors. Four pattern recognition methods were employed: support vector machines (SVM), multilayer perceptron (MLP), nearest neighbors (KNN) and naive Bayes (NBY). These four methods were chosen  because they usually display a good overall performance~\cite{systematic}. An introduction to these methods can be found in~\cite{systematic,bishop}. We also provide a very short introduction to these methods in the Supplementary Information\footnote{The Supplementary Information is available from \url{https://dl.dropboxusercontent.com/u/2740286/symmetry.pdf}}.
%


\section{Results and discussion}

This section is divided in two subsections. Firstly, we study the statistical properties of symmetry measurements in word adjacency networks. We then show how the symmetry of specific words can be employed to discriminate authors' styles. The list of books employed in the experiments is shown in Table S1 of the {Supplementary Information}.

\subsection{Properties of merged and backbone symmetry in word adjacency networks}

\begin{figure}[!htb]
\onefigure[width=1.0\linewidth]{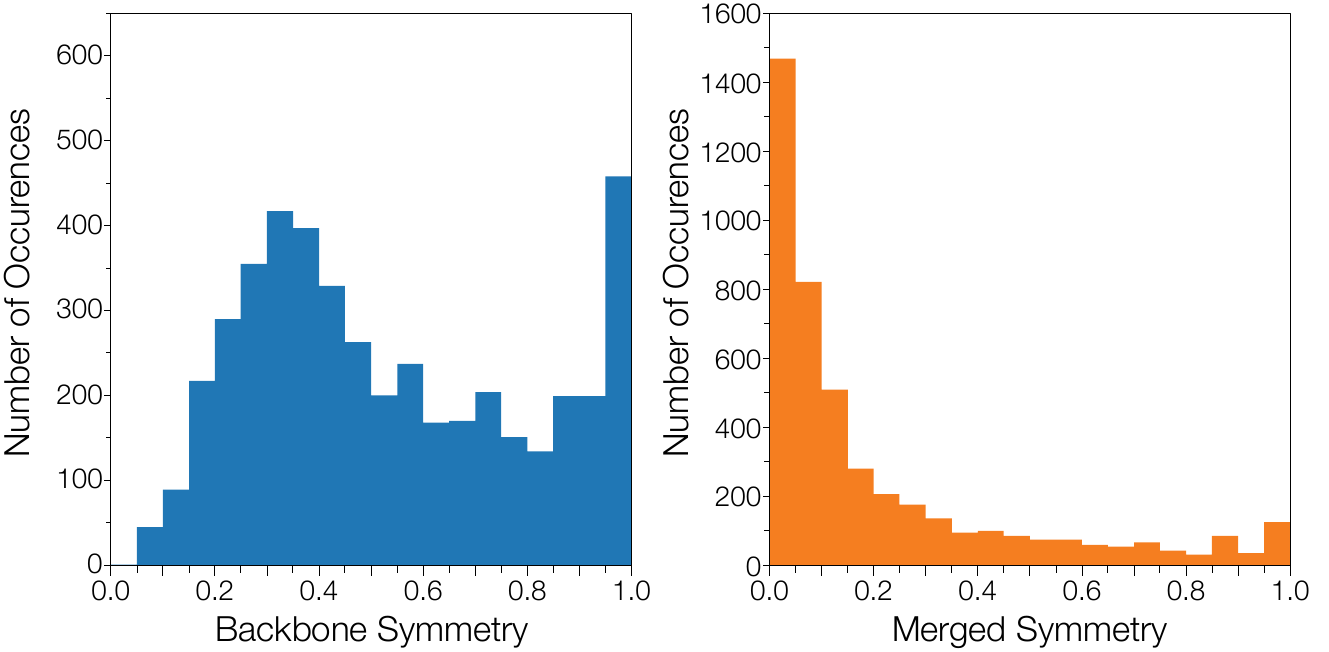}
\caption{Histograms of the distribution of the backbone $S_b$ and merged symmetries $S_m$ (computed at the second level) for the book ``Adventures of Sally''. The merged symmetry computed at the second level seems to follow a logistic function (see equations \ref{log1} and \ref{log2}). A similar distribution was found for the other books of the dataset.}
\label{fig.1}
\end{figure}

We start the investigation of the statistical properties of symmetry measurements in textual networks by analyzing the distribution of symmetry values in real networks formed from books. Here we focus our discussion on the book ``Adventures of Sally'', by P.G. Wodehouse. Notwithstanding, all discussion henceforth applies to the other books of the dataset. Concerning the merged symmetry, all books displayed a probability density function with the following logistic form:
\begin{equation} \label{log1}
    P(S_m) \simeq  \frac{A_1 - A_2}{1+(S_m/S_0)^p} + A_2,
\end{equation}
where $A_1$, $A_2$, $S_0$ and $p$ are constant. According to the equation \ref{log1}, high values of symmetry are very rare. This is similar to other well-known distributions in texts, such as the frequency distribution given by the Zipf's law~\cite{origin}. Fig. \ref{fig.1} illustrates the histogram of symmetry distribution obtained for the book ``Adventures of Sally'', by P.W. Wodehouse. For this book in particular, the p.d.f of the merged symmetry in equation \ref{log1} can be written as
\begin{equation} \label{log2}
    P(S_m) \simeq \frac{A}{ 1 + (S_m / S_0)^{p}},%
\end{equation}
where $A = 1.0136$, $S_0 = 0.0136$ and $p=1.25348$. The high value of adjusted Pearson ($R^2 = 0.99181$) and low value of chi-square ($\chi^2 = 1.43261 \cdot 10^{-5}$) confirm the adehenrece of the fitting.

Unlike the merged symmetry, the backbone counterpart displayed a distribution of values with two typical peaks, as revealed by Fig.~\ref{fig.1} (see left panel). The first peak of distribution occurs around $S_b \simeq 0.3$. While low values of backbone symmetry are very rare, high values are frequent, especially on the less frequent words. This occurs because smaller (or lowly connected) concentric patterns are more unlikely to accumulate enough imperfections over the concentric levels to attain very low symmetry values. On the other hand, larger patterns do not present such constraints and can attain many distinct levels of symmetry.

%
%
%

While several traditional centrality network measurements correlate with the node degree, the proposed symmetry measurements for text analysis usually do not yield a strong correlation with the connectivity of nodes. In Tables \ref{tab.00} and \ref{tab.01}, we show, in the same row, words with similar degree taking very discrepant values of merged and backbone symmetries. For example, in Table \ref{tab.01}, the words \emph{bathing} and \emph{mother} occur with the same frequency; however, the respective values of backbone symmetry are quite discrepant. As a matter of fact, the access to the second level neighbors is much more regular for the word \emph{mother}, as it backbone symmetry is close to the maximum possible value, i.e. $\max (S_b) = 1$.
\begin{table}[h]
    \caption{Merged symmetry (second level) computed for selected words in the book ``Adventures of Sally'', a novel by P.G. Wodehouse. Note that words with similar degree $k$ (the words in the same line) may take distinct values of symmetry.}
    \label{tab.00}
    \begin{center}
        \begin{tabular}{|l|cc|l|cc|}
            \hline
            {\bf Word}   &  $S_m$ & $k$ & {\bf Word} & $S_m$ & $k$ \\
            \hline
             Cracknell &   0.011 & 31 & hotel        & 0.024 & 33 \\
             heart     &   0.012 & 27 & corner       & 0.029 & 27 \\
             gentleman &   0.012 & 26 & conversation & 0.041 & 24 \\
             revue     &   0.012 & 21 & rise         & 0.062 & 21 \\
             notice    &   0.013 & 17 & blow         & 0.080 & 17 \\
             cold      &   0.014 & 11 & tongue       & 0.094 & 10 \\
             luck      &   0.017 & 6  & wealth       & 0.331 & 6  \\
             meditate  &   0.020 & 5  & banquet      & 0.259 & 5  \\
            \hline
        \end{tabular}
    \end{center}
\end{table}
\begin{table}[h]
    \caption{Backbone symmetry (second level) computed for selected words in the book ``Adventures of Sally'', a novel by P.G. Wodehouse. Note that words with similar degree $k$ (the words in the same line) may take distinct values of symmetry.}
    \label{tab.01}
    \begin{center}
        \begin{tabular}{|l|cc|l|cc|}
            \hline
            {\bf Word}   &  $S_b$ & $k$ & {\bf Word} & $S_b$ & $k$ \\
            \hline
             hair      &   0.196 & 30 & hotel        & 0.472 & 33 \\
             heart     &   0.211 & 27 & corner       & 0.412 & 27 \\
             manner    &   0.190 & 26 & conversation & 0.544 & 24 \\
             chapter   &   0.127 & 22 & york         & 0.579 & 19 \\
             water     &   0.132 & 16 & disappear    & 0.610 & 14 \\
             note      &   0.052 & 10 & mysterious   & 0.709 & 8  \\
             memory    &   0.089 & 6  & secure       & 0.904 & 6  \\
             bathing   &   0.071 & 5  & mother       & 0.932 & 5  \\
            \hline
        \end{tabular}
    \end{center}
\end{table}

\begin{figure}[!htb]
\onefigure[width=0.90\linewidth]{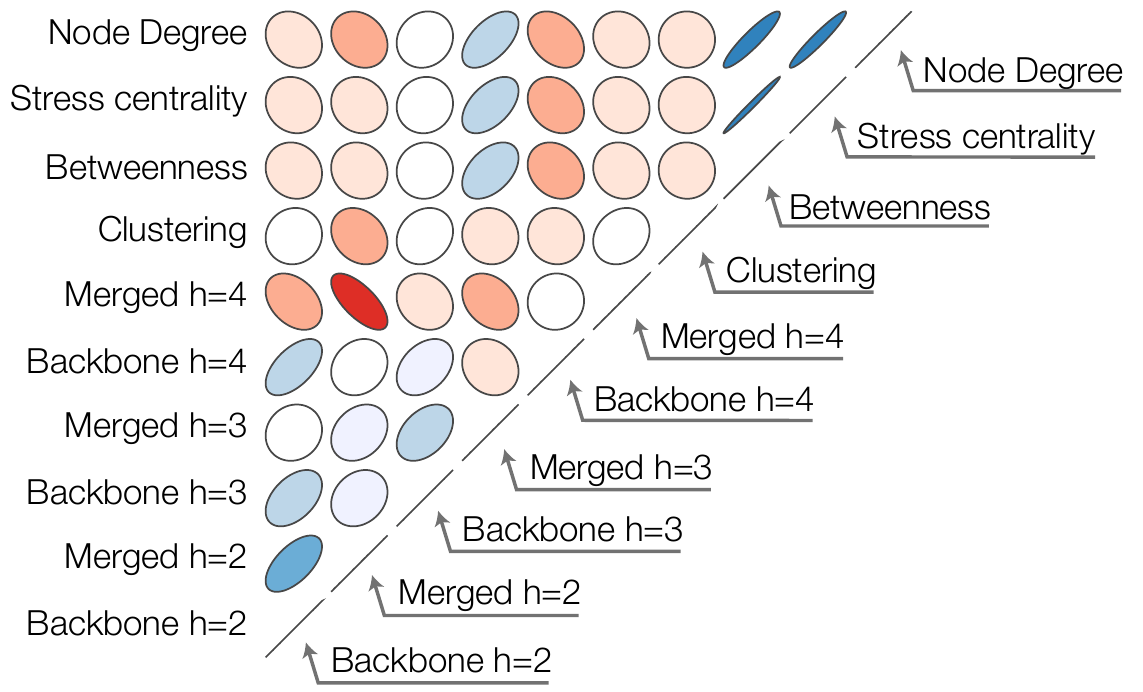}
\caption{Pearson correlation coefficients between symmetry and other traditional network measurements. Note that, in general, there is a weak correlation between symmetry and other measurements. The correlations were obtained from the word adjacency network obtained from the book ``Adventures of Sally'', by P.G. Wodehouse.}
\label{fig.corr}
\end{figure}

The correlation between symmetry and other traditional topological measurements were also investigated. According to Fig. \ref{fig.corr}, there is no consistent, significant correlation between symmetry and other network measurements. This means  that the values of both merged and backbone symmetry cannot be mimicked by other well known network measurements. Therefore, the symmetry measurements provide novel information for network analysis.
%

\subsection{Authorship recognition via network symmetry}

In this section, we exemplify the discriminability power of symmetry measurements in word adjacency networks. More specifically, we show that the symmetry os specific words is able to identify the writing style of distinct authors.
In the context of information sciences, the authorship recognition task is relevant because it can be useful to classify literary manuscripts~\cite{74} and intercept terrorist messages~\cite{73}. Traditional features employed for stylometric analysis include simple statistics such as the average length and frequency of words~\cite{49}, richness of vocabulary size~\cite{49} and burstiness indexes~\cite{nova}.

To evaluate the ability of the symmetry measurements to recognize particular authors' styles, we used a dataset of 40 books written by 8 authors {(see Table S1 of the Supplementary Information)}. As features for the classification task, both merged and backbone symmetry were computed for the $229$ words appearing in all books of the dataset. To automatically recognize and classify the patterns displayed by each author, we used the four pattern recognition techniques described in the methodology. The accuracy rates in identifying the correct author are shown in Table~\ref{tab.acc}. With regard to the performance of the pattern recognition methods, the best results were obtained with the SVM and MLP methods. When the symmetry was computed considering the second level of neighbors ($h=2$), the best accuracy rate achieved was 75.0\% (this corresponds to a $p$-value lower than $1.0 \cdot 10^{-15}$). Both symmetries measurements calculated at the third level did not increase the best classification performance obtained with $h=2$. A minor improvement in performance occurred when the fourth level was included in the analysis. The best accuracy rate increased from 75.0\% to 82.5\%. We also probed the performance of the classification by combining different levels as features. In this case, the performance did not improve (result not shown). All in all, these results confirms the suitability of symmetry measurements to identify the subtleties of authors' styles in terms of the homogeneity of accessibility of neighbors.

\begin{table}[h]
    \caption{Accuracy rate found for the authorship recognition task. The best accuracy rate found to recognize the authorship in a dataset comprising 8 authors was 82.5\%. }
    \label{tab.acc}
    \begin{center}
        \begin{tabular}{|l|cccc|}
            \hline
            {\bf Symmetry} & {\bf SVM} & {\bf MLP} & {\bf KNN} & {\bf NBY} \\
            \hline
            Merged   $h=2$   & 75.0\% & 72.5\% & 55.0\% & 42.5\%  \\
            Merged   $h=3$   & 70.0\% & 62.5\% & 65.0\% & 40.0\%  \\
            Merged   $h=4$   & 82.5\% & 82.5\% & 57.5\% & 42.5\%  \\
            Backbone $h=2$ & 32.5\% & 32.5\% & 20.0\% & 20.0\%  \\
            Backbone $h=3$ & 70.0\% & 72.5\% & 57.5\% & 27.5\%  \\
            Backbone $h=4$ & 70.0\% & 82.5\% & 57.5\% & 42.5\%  \\
            \hline
        \end{tabular}
    \end{center}
\end{table}

\begin{figure*}[!htb]
\onefigure[width=17.5cm]{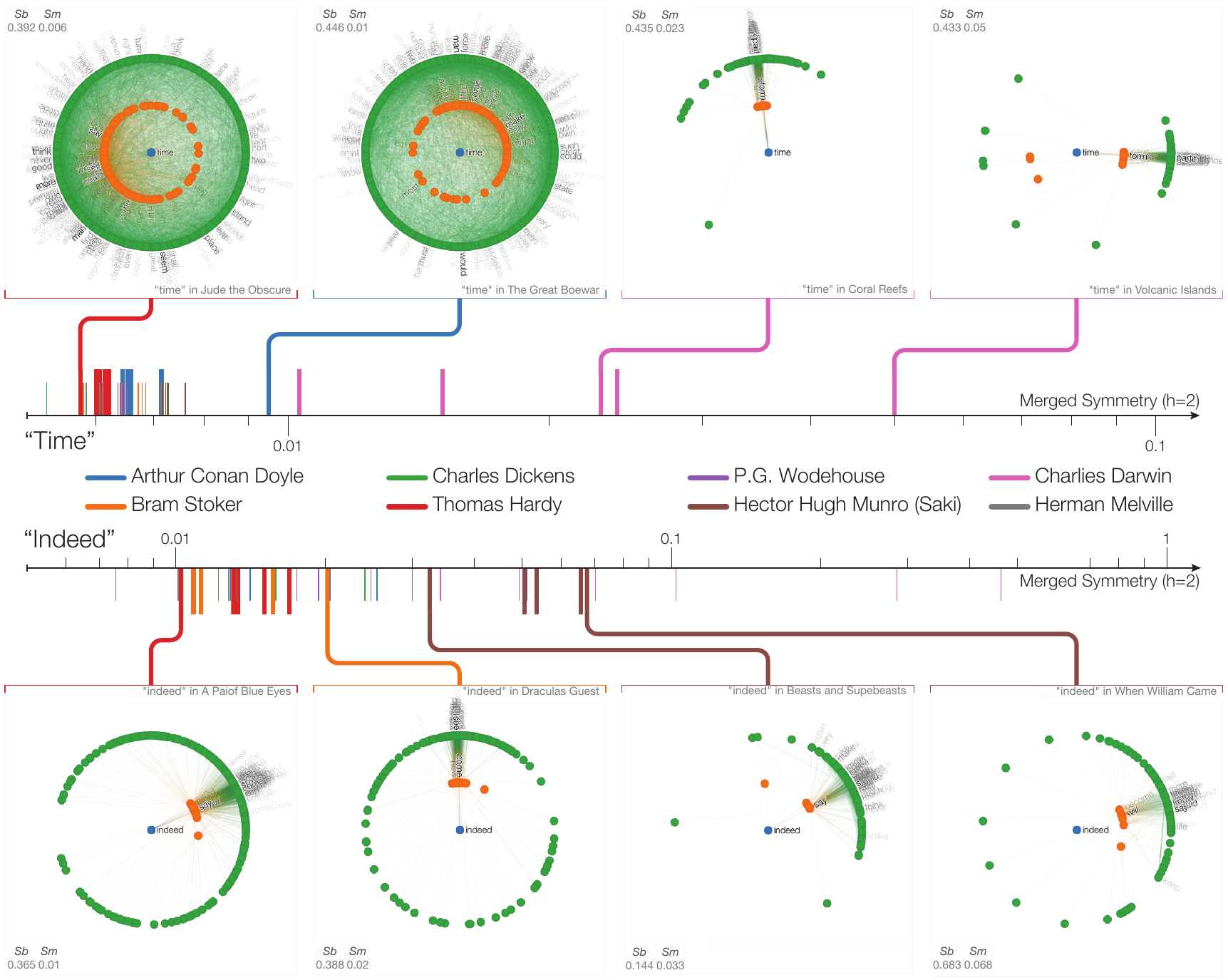}
\caption{Merged symmetry values for considered books and visualizations of a few concentric patterns obtained by considering the words ``time'' and ``indeed''. Each bar over the axis correspond to a book and the colors indicate their respective authors according to the legend. The patterns visualizations was accomplished by using a modified force directed method~\cite{force} where nodes connected in the same level are more likely to be close together. Words shared among all books are also shown next to the respective nodes with the opacity proportional to its frequency.}
\label{fig:time_line}
\end{figure*}

To understand the patterns behind the high discriminability rates found in Table~\ref{tab.acc} we show some visualizations obtained for two words, ``time'' and ``indeed'', in Fig.~\ref{fig:time_line}. We chose these words because they were able discriminate among a few groups of authors while also presenting a wide range of symmetry values. The patterns obtained for the word ``time'' are arranged along the top of the corresponding axis according to their respective merged symmetry, which was found to separate \emph{Arthur Conan Doyle}, \emph{Thomas Hardy} and \emph{Charles Darwin}. Note that the nodes with low merged symmetry presented several edges crossing over the internal shell of its patterns. Additionally, connections between nodes lying at the third concentric level are much less organized, hence the low values of symmetry. Conversely, nodes taking high values of symmetry displayed more organized connections, leading to higher uniformity of connections among nodes lying at the farthest concentric level. The same observations can be made for the patterns obtained for the word ``indeed'', which discriminated between \emph{Hector Hugh Munro} and the group of authors encompassing \emph{Arthur Conan Doyle}, \emph{Bram Stoker}, \emph{Thomas Hardy} and \emph{Charles Dickens}. Still, however, these patterns are much more symmetric, which is once again reflected in the visualizations by their higher organization on the last concentric level.



\section{Conclusion}

In this paper, we have introduced the concept of symmetry to study the connectivity patterns of word association networks. By defining symmetry as a function of particular random walks, we showed that the symmetry measurements are robust in the sense that they do not mimic the behavior of other traditional topological measurements.
Thus, because symmetry measurements do not strongly correlate with other traditional network or textual features, they could be combined with other measurements to improve the characterization of texts represented as graphs and related networked systems.
The proposed symmetry measurements were also evaluated in the context of the authorship recognition task. The results revealed that the symmetry of specific words is able to identify the authorship of books with high accuracy rates. This result confirms the suitability of the measurements to detect the subtleties  of authors' styles reflected on the organization of word adjacency networks. In future works, we intend to study the suitability of both backbone and merged symmetry in semantical networks, which may ultimately lead to the improvement of several semantical-related applications.
%
%
%





\acknowledgments
DRA acknowledges financial support from S\~ao Paulo Research Foundation (FAPESP) (grant number 2014/20830-0). FNS thanks CAPES for support. LdFC is grateful to FAPESP (grant number 2011/50761-2), CNPq (Brazil) and NAP-PRP-USP.


\begin{thebibliography}{0}

\bibitem{surveyApp}
\Name{Costa L.F. et al.}
\REVIEW{Adv. Phys.}{60}{2011}{329--412}.
%

\bibitem{cong}
\Name{Cong J. \and Liu H.}
\REVIEW{Phys. Life Rev.}{11}{2014}{598--618}.

\bibitem{extractive}
\Name{Amancio D.R., Nunes M.G.V., Oliveira Jr. O.N. \and Costa L.F.}
\REVIEW{Physica A}{391}{2012}{1855--1864}.

\bibitem{wsd}
\Name{Amancio D.R., Oliveira Jr. O.N. \and Costa L.F.}
\REVIEW{Europhys. Lett.}{98}{2012}{18002}

\bibitem{compSanda}
\Name{Amancio D.R., Aluisio S.M., Oliveira Jr. O.N. \and Costa L.F.}
\REVIEW{Europhys. Lett.}{100}{2012}{58002}


\bibitem{ortuno}
\Name{Ortu\~no M., Carpena P., Bernaola-Galvan P., Mu\~noz E. \and Somoza A.M.}
\REVIEW{Europhys. Lett.}{57}{2002}{759}

\bibitem{nova}
\Name{Amancio D.R.}
\REVIEW{J. Stat. Mech.}{}{2015}{P03005}.

\bibitem{patterns}
  \Name{Ferrer i Cancho R., Sol\'e R.V. \and Kohler R.}
  \REVIEW{Phys. Rev. E}{69}{2004}{1--8}.

\bibitem{literary}
\Name{Amancio D.R., Oliveira Jr. O.N. \and Costa L.F.}
\REVIEW{New J. Phys.}{14}{2012}{043029}.

\bibitem{concentric}
\Name{Costa, L.da F., Tognetti, M.A.R. \and Silva, F.N.}
\REVIEW{Physica A}{24(387)}{2008}{6201-6214}.

\bibitem{short}
\Name{Amancio D.R.}
\REVIEW{PLoS ONE}{10}{2015}{e0118394}.

\bibitem{poetry}
\Name{Roxas R.M. \and Tapang G.}
\REVIEW{Int. J. Mod. Phys. C}{21}{2010}{503}.

\bibitem{voynich}
\Name{Amancio D.R., Altmann E.G., Rybski D., Oliveira Jr. O.N. \and Costa L.F.}
\REVIEW{PLoS ONE}{8}{2013}{e67310}.

\bibitem{statnlp}
\Name{Manning C.D. \and Schutze H.}
\Book{Foundations of Statistical Natural Language Processing}
\Publ{MIT Press}
\Year{1999}.

\bibitem{ratna}
\Name{Ratnaparki A.}
\Book{Proceedings of the Empirical Methods in Natural Language Processing Conference}
\Year{1996}.



\bibitem{sym1}
\Name{Debs T. \and Redhead M.}
\Book{Objectivity, Invariance, and Convention: Symmetry in Physical Science}
\Publ{Harvard Univ. Press}
\Year{2007}

\bibitem{sym2}
\Name{MacArthur B.D. \and Anderson J.W.}
\REVIEW{arXiv: cond-mat/0609274}{}{2006}{}

\bibitem{sym3}
\Name{Finnerty J.R.}
\REVIEW{Int. J. Dev. Biol.}{47}{2003}{523–9}

\bibitem{sym4}
\Name{Longuet-Higgins H.C.}
\REVIEW{Mol. Phys.}{6:5}{1963}{445--460}

\bibitem{holme}
\Name{Holmes P.}
\REVIEW{Phys. Rev. E}{74}{2006}{036107}

\bibitem{ingleses}
\Name{Rossi L., Torsello A., Hancock E.R. \and Wilson R.C.}
\REVIEW{Phys. Rev. E}{88}{2013}{032806}

\bibitem{mir}
\Name{Silva F.N.,  Comin C.H., Peron T.K.D., Rodrigues F.A., Ye C., Wilson R.C., Hancock E. \and Costa L.F.}
\REVIEW{arXiv: 1407.0224}{}{2014}{}

\bibitem{viana}
\Name{Viana M.P., Batista J.L.B. \and Costa L.F.}
\REVIEW{Phys. Rev. E}{85}{2012}{036105}

\bibitem{duda}
\Name{Duda R.O., Hart P.E. \and Stork D.G.}
\Book{Pattern Classification}
\Vol{2}
\Publ{Wiley-Interscience}
\Year{2000}


\bibitem{systematic}
\Name{Amancio D.R., Comin C.H., Casanova D., Travieso G., Bruno O.M., Rodrigues F.A. \and Costa L.F.}
\REVIEW{PLoS ONE}{9}{2014}{e94137}.

\bibitem{bishop}
\Name{Bishop C.M.}
\Book{Pattern Recognition and Machine Learning}
\Publ{Springer-Verlag New York, Inc. Secaucus, NJ, USA}
\Year{2006}.

\bibitem{origin}
\Name{Zipf G.K.}
\Book{Human behavior and the principle of least effort}
\Publ{Addison-Wesley, Reading MA, USA}
\Year{1949}

\bibitem{74}
\Name{Ebrahimpour M., Putnins T.J., Berryman M.J., Allison A., Ng BW-H. \and Derek A.}
\REVIEW{PLoS ONE }{8}{2013}{e54998}.

\bibitem{73}
\Name{Abbasi A. \and Chen H. }
\REVIEW{IEEE Intell. Syst.}{20}{2005}{67--75}.



\bibitem{49}
\Name{Stamatatos E.}
\REVIEW{J. Assoc. Inf. Sci. Technol.}{60}{2009}{538--556}.

\bibitem{force}
\Name{Fruchterman T.M.J. \and Reingold E.M.}
\REVIEW{Softw: Pract. Exper.}{11(21)}{1991}{1129--1164}

\end{thebibliography}
\end{document}